# Comparative Evaluation of Tree-Based Ensemble Algorithms for Short-Term Travel Time Prediction

**Saleh Mousa,**

Research Assistant

Department of Civil and Environmental Engineering

Louisiana State University, Baton Rouge, LA 70803, USA

Email: smousa6@lsu.edu

**Sherif Ishak, Ph.D, P.E, Professor**

Department of Civil and Environmental Engineering

Old Dominion University

E-mail address: sishak@odu.edu

**ABSTRACT**

Disseminating accurate travel time information to road users helps achieve traffic equilibrium and reduce traffic congestion. The deployment of Connected Vehicles technology will provide unique opportunities for the implementation of travel time prediction models. The aim of this study is twofold: (1) estimate travel times in the freeway network at five-minute intervals using Basic Safety Messages (BSM); (2) develop an eXtreme Gradient Boosting (XGB) model for short-term travel time prediction on freeways. The XGB tree-based ensemble prediction model is evaluated against common tree-based ensemble algorithms and the evaluations are performed at five-minute intervals over a 30-minute horizon. BSMs generated by the Safety Pilot Model Deployment conducted in Ann Arbor, Michigan, were used. Nearly two billion messages were processed for providing travel time estimates for the entire freeway network. A Combination of grid search and five-fold cross-validation techniques using the travel time estimates were used for developing the prediction models and tuning their parameters. About 9.6 km freeway stretch was used for evaluating the XGB together with the most common tree-based ensemble algorithms. The results show that XGB is superior to all other algorithms, followed by the Gradient Boosting. XGB travel time predictions were accurate and consistent with variations during peak periods, with mean absolute percentage error in prediction about 5.9% and 7.8% for 5-minute and 30-minute horizons, respectively. Additionally, through applying the developed models to another 4.7 km stretch along the eastbound segment of M-14, the XGB demonstrated its considerable advantages in travel time prediction during congested and uncongested conditions.

## 1. INTRODUCTION

According to the 2015 Annual Urban Mobility Report, the yearly delay per auto commuter increased from 37 hours in year 2000 to 42 hours in 2014 [1]. Likewise, the total annual congestion cost, resulting from travel delays and wasted fuel by all vehicles in the US, has increased from $114 billion in year 2000 to $160 billion in 2014 [1]. These alarming indicators raise the importance of sharing reliable travel time estimation and prediction values with road users on real time basis. Travel time is an effective measure of traffic conditions and reflects mobility on the road network, which is easily accepted by travelers more than any other measure such as queue length. Providing reliable travel time predictions through Advanced

Traveler Information Systems (ATIS) enables travelers to avoid commuting along links experiencing high travel times during peak periods. This can be achieved by either rerouting or rescheduling their trips for later times which would benefit travelers and provide traffic equilibrium with less congested traffic streams.

With the rapid development of information and communication technology, equipped automobiles with wireless communication capabilities are expected to be the next frontier for automotive revolution. This will accelerate the deployment of fully connected transportation systems in order to provide safer roads, improved mobility, and green environment [2, 3]. The market of connected vehicles (CV) is booming, and according to recent business reports, the global market is expected to reach $131.9 billion by 2019 [4]. The CV provide opportunities to expand the geographical scope and enhance the precision of Traffic Management Centers (TMC) controlling the transportation network. As a result, CV data are processed in this study to build a spatial-temporal travel time matrix for the entire freeway network. However, this raises the need for receiving, analyzing, and processing a large amount of data even at low market penetration values. For instance, one of the largest datasets collected by CV, used in this study, is the Safety Pilot Model Deployment (SPMD) conducted in Ann Arbor, Michigan from August 2012 to late 2013. During this period, the SPMD generated over one terabyte of data per month from only 2,836 vehicles [5,6]. Basic Safety Messages (BSM), known globally as messages containing vehicle safety information and transmitted periodically to surrounding vehicles, are a subset of the SPMD dataset. Adopting such CV data means the need for tools to analyze and process large data at TMCs on a real-time basis. This shifted the global trend towards big data analytics and High Supercomputers for utilizing CV data in traffic predictions. However, a more sustainable and efficient approach for estimating traffic measures from CV environment would be through allowing the TMC to use CV as processing nodes based on their geographic location for aggregating the BSM within a margin of this location.

To this end, the first objective of this paper is performing travel time estimation along the freeway network using BSM. Travel time estimation is performed for all freeway segments using six traditional computers, simulating the possibility of processing the massive data using six vehicles. Utilizing the SPMD data for estimating traffic measures is documented in the literature in studies with different scopes, the most relevant to the scope of this study is congestion identification and prediction [7].

The short-term traffic state prediction is not a straightforward task because of the instability of traffic conditions and complex road settings [8]. During the past decades, different studies with various approaches have been conducted for short-term traffic state prediction. Technical details of these approaches can be found in these references [9, 10]. Methods used include time series, regression, and stochastic approaches [11, 12, 13, 14]. The computational advances in traditional processing powers catalyzed the evolution of machine learning algorithms in this area. Inspired by their data-driven nature, these algorithms do not assume any preliminary model structure for data, giving flexibility for processing complex data. Some of these machine learning prediction models include neural networks [15, 16], support vector regression (SVR) [17, 18], and genetic algorithms [19, 20]. Other hybrid algorithms were also used, incorporating performing cluster analysis prior applying the machine-learning algorithm for prediction [21, 22].

In recent years, ensemble-based algorithms became more popular and convenient in solving regression and classification problems. These algorithms combine predictions of several base estimators built with a given learning algorithm to improve robustness and accuracy over a single estimator. Among the vast diversity of ensemble methods, the tree-based ensemble algorithms are the most common ones. Instead of fitting the best single decision tree, the model combines multiple trees to enhance prediction accuracy. In fact, they can model complex nonlinear relationships [23]. Unlike most machine learning tools, the possible interpretability of tree-based ensemble models makes them a good candidate for solving prediction problems incorporating a large number of input variables. However, the literature lacks studies implementing ensemble methods for traffic state prediction. Far to the knowledge of the author, the literature includes only three travel time prediction studies implementing tree-based ensemble methods. Two studies implemented Random Forests (RF) [24, 25] and the third implemented Gradient Boosting (GB) algorithms [26] for travel time prediction.

In 2010, Kaggle was founded as a platform for predictive modeling and analytics competitions where companies and researchers post their data and compete for producing the best models [27]. The eXtreme Gradient Boosting (XGB) algorithm, developed recently, is considered one of the most accurate algorithms. Since the development of the XGB, it has won most competitions in structural data category, including Kaggle. Although XGB shares a common base algorithm with GB approach, yet it is a more regularized model formalization, that controls modeling the noise within the data (prevents overfitting). This makes XGB outperform all other tree-based ensemble algorithms [28]. The efficiency, accuracy, feasibility, and short processing time are additional advantages of the XGB model over common models such as RF and Neural Networks. This puts the XGB as a valuable tool in solving transportation application problems including travel time prediction.

The review of the literature revealed a lack of research studies on the implementation of XGB algorithms to solve traffic state prediction problems. To this end, as a second and main objective, this paper develops an XGB short-term travel time prediction model for freeways. The XGB prediction model is evaluated against the most common tree-based ensemble algorithms and the evaluations are performed at five-minute intervals over a 30-minute horizon. In this study, the travel time estimation is first performed for all freeway segments, and then, the tree-based travel time prediction models, including XGB, are developed. The XGB model is evaluated against all common ensemble tree-based models and a single decision tree model as well. Moreover, the developed models are applied over a 4.7 km stretch along the eastbound of M-14 to assess the performance of the models during both congested and uncongested conditions. A description of the utilized tree-based ensemble algorithms is highlighted in the next section, followed by an outline of the of the travel time estimation algorithm using BSM. Then, the analysis performed for building the prediction models is summarized. Finally, the paper presents the analysis of results, the features of the best prediction model, and the overall findings of the study.

## 2. TECHNICAL BACKGROUND

This section provides technical background for the single decision tree model, followed by a description for the utilized tree-based ensemble models.

### 2.1. Single Decision Tree

Decision tree (DT) is a data-driven supervised learning method, which builds a set of decision rules from all input variables to predict response variable. Decision rules are presented as nodes, splitting features space into regions or sub-nodes. Each sub-node is further split until a specific criterion is met. Each terminal node of these structures is called leaf and is assigned a constant score value (*C*), which is typically the average of response variables in this node. In this essence, a tree can be defined as a vector of leaf scores and a leaf index mapping function. For a given data set with (*n*) observations and (*m*) input variables, the general formulation of this structure can be presented as follows:

$$f\,(x) = C_{q(x)}, (q \colon \mathbb{R}^m \to \{1,2,..,T\}, C \in \mathbb{R}^m) \tag{1}$$

Where, $q_{(x)}$ represents the decision rules within a tree that assigns sample of the data to the corresponding leaf index, (T) is the number of leaves in the tree, and $C_{q(x)}$ represents the score weights assigned to leaves of the tree.

### 2.2. Tree-Based Ensemble Methods

All ensemble tree-based algorithms consist of base multiple models (trees) that are combined together to enhance prediction accuracy. As a result, a general prediction model (ŷ) can be written for ensemble models as an aggregation of all prediction scores for all trees for a sample (*x*). The general formulation of these models is given in below in conjunction with Eq. (1):

$$\hat{y}_i\,(x) = \sum_{k=1}^{K} f_k\,(x_i), \qquad (f_k \in \mathcal{F}) \tag{2}$$

Where, (*k*) is the number of trees and ($\mathcal{F}$) is the space of all possible regression trees. This equation is to be optimized for the following objective function:

$$Obj(\Theta) = \sum_{i=1}^{n} l(y_i, \hat{y}_i) + \sum_{k=1}^{K} \Omega(f_k) \tag{3}$$

Where the first term is the loss function measuring the difference between prediction ($\hat{y}_i$) and target($y_i$). The second term is the regularization term controlling the model complexity and preventing overfitting. All ensemble tree based algorithms have the same general model; the only difference is how training is done and regularization is accounted for as presented in the following subsections.

*1) Bagging Algorithms*

This family of ensemble models uses bootstrap samples to build multiple trees independent of the original training set and then report the average of trees. This procedure of random sampling with replacement reduces the variance of a single decision tree and can add substantial improvement in the performance over unstable learners [29].

*2) RF Algorithms*

The random sampling with replacement in large data sets can grow similar trees in the bagging algorithm. For this purpose, RF does not consider all features at each split and uses a different random subset of features during splitting for each tree. Such adjustment reduces the correlation among trees and never builds similar trees, which enhances prediction accuracy.

*3) Adaboost Algorithms*
In this family of algorithms, trees are grown in sequential order, and the final predictor is a weighted sum of each individual tree. The core principle of the Adaboost algorithm is applying weights to each training sample and updating these weights in each boosting iteration based on the prediction accuracy of that specific sample in the previous iteration. As iterations proceed, cases that were predicted with less accuracy will receive higher weights while lower weights are assigned to those predicted correctly. In this manner, preceding learners will focus on observations with low accuracy. Finally, predictions from all trees are combined through weighted sum to produce the final prediction.

*4) GB Algorithms*
Similar to Adaboost, GB follows the same fundamental approach; where trees in this algorithm are grown sequentially to improve the robustness of the algorithm against overlapping class distributions by optimizing an arbitrary differentiable loss function using Gradient Descent method [30]. In summary, the main idea of GB is to sequentially fit different tree at each iteration, which gradually minimizes the loss function of the whole system and not to re-weighted observations, as in Adaboost [30].

*5) XGB Algorithms*
*XGB* is an advanced implementation of GB, which received wide recognition among machine learning community due to its accuracy, efficiency, and ease of implementation. Unlike the previous boosting algorithms (Adaboost and GB) which account for regularization in the model by only considering a learning rate (*L*), which shrinks the contribution of each successive tree, *XGB* implements an additional regularization term as shown below:

$$\Omega(f_k) = \gamma T + \frac{1}{2}\lambda \sum_{j=1}^{T} C^2{}_j \tag{4}$$

This regularization term penalizes complicated models on the number of leaves and the sum of squared scores of leaves. This means that simpler models are preferred over complex ones. The $\gamma$ and $\lambda$ are two regularization parameters. A comprehensive solution for how to solve (1), (2), (3) and (4) for training the XGB algorithm can be found in this study [31]. Another remarkable features of the XGB are the cache-line and memory optimizations performed prior training the data, making the training patterns faster and cache friendly. As a result, this recently developed algorithm outperforms the GB in the scalability and processing speed with relatively higher accuracy [32, 33].

## 3. TRAVEL TIMEESTIMATION USING BSM
This section discusses the steps performed for estimating travel times using BSM. It also describes the study area, BSM data, the algorithm for processing, and preparing data for travel time prediction.

### 3.1. BSM and Study Area
This study used the BSM data set of the SPMD during April 2013 to estimate travel times for the whole month. BSM is available for public on the USDOT Research Data Exchange (RDE) website [5]. The data contains a timestamp at 10 Hz frequency for each vehicle. This stamp reports, for a specific coded time, the location of the vehicle (longitude and latitude), direction (heading), dynamic state (longitudinal acceleration, lateral acceleration, speed, and yaw rate), and other information. Each stamp includes a total 19 variables, out of which only four (time, longitude, latitude, and speed) are needed for travel time estimation. However, the longitudinal and lateral accelerations were included in the analysis as well, in case they are needed in the prediction process. Estimating travel time from this vast amount of data at reasonable durations is a challenge because the data includes all time stamps reported everywhere in the network (freeways, arterial, collector, etc.). In order to estimate travel times for all freeways surrounding Ann Arbor City, the freeways were divided into 28 sections based on a maximum threshold length of 2 km, curvature, and exit/entry ramps, and were geocoded in an ArcGIS map as shown in Figure 1.

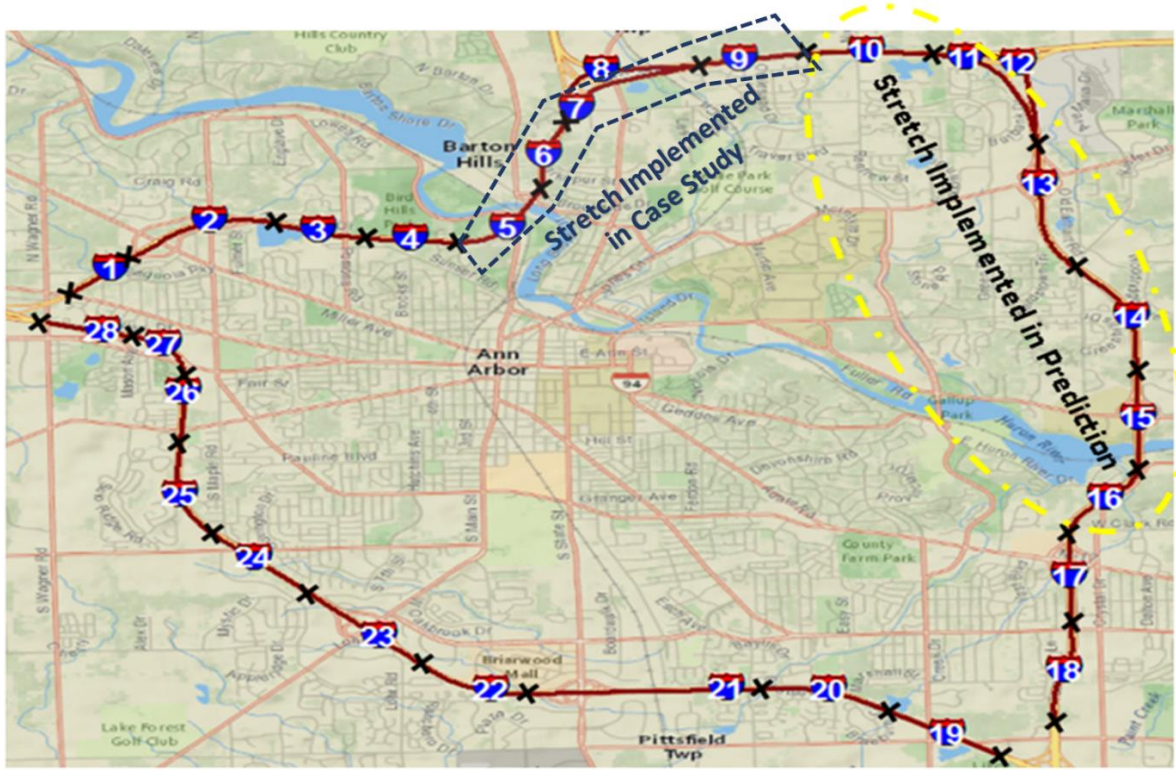

**Figure 1: Layout of Freeway Segments Bounding Ann Arbor City**

### 3.2. Travel Time Estimation Algorithm

The logic of the travel time estimation algorithm is illustrated in Figure 2. First, the one-month data were reduced by excluding 13 irrelevant variables, leaving only six variables relevant to travel time prediction. Then, the reduced data were divided into smaller files such that each file can be handled and processed with a traditional computer without running out of the memory (510 files in this study).

The needed geospatial analysis and decoding of the time were performed in ArcGIS to extract the stamps associated with each link. Based on these stamps, the speed and the lateral and longitudinal accelerations were aggregated for each link, by calculating the average values at 5-minute intervals, and the heading was used to identify opposite directions. Finally, the segment length was used to estimate travel times for each time step. To this end, a python code was written and implemented in ArcGIS environment. In addition, this code distributed the processing tasks across six traditional computers.

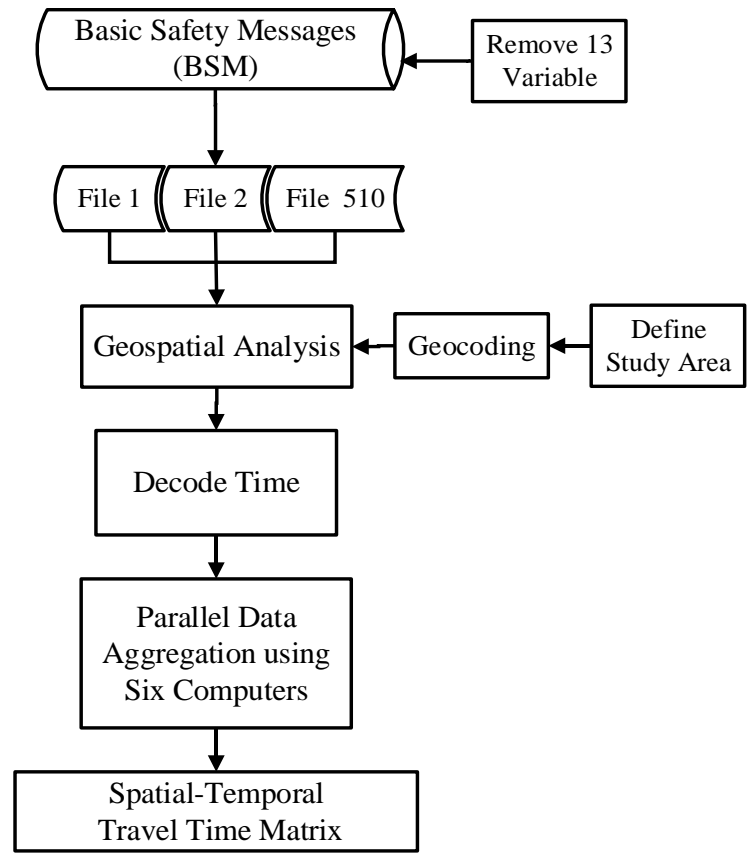

**Figure 2: Logic of the Travel Time Estimation Algorithm**

In this research, the distribution of data among the six computers was based on the geographical location of the message. This approach simulates the deployment of six connected vehicles or six roadside units in the network for processing data and providing travel time estimates to the TMC. In reality, the TMC can coordinate collecting processed aggregated travel times from different vehicles/road site units based on their location and develop the final spatial-temporal travel time matrix for the entire network, instead of processing the entire data. The gains from such approach are marginal when estimating travel times for training the prediction model, where the model training can be done offline and no need for real-time processing. However, when applying the prediction model for a specific time step, all stamps generated during this five-minute duration step will need to be mapped and aggregated on a real-time basis and the processing time gains can be substantial.

### *3.3. Preparing the Data for Travel Time Prediction*

A stretch consisting of six links along the southbound of US-13 was used for travel time prediction. This stretch is about 9.6 km long and consists of links no. 10, 11, 13, 14, 15 and 16 as shown in Figure. 1. These links are referred to as 1 through 6, respectively. After performing interpolations for intervals with missing values, the total number of time steps available for prediction was 8,640. Traffic volumes counted at stations located on US-13 were below 1000 veh/hr/lane between 8:00 PM - 5:00 AM with almost constant travel times. This period was excluded and only intervals between 5:00 AM and 8:00 PM were considered with 5,400 steps.

## 4. DEVELOPING TRAVEL TIME PREDICTION MODEL

Selecting the variables input to the prediction model is an essential task, due to the interrelation between the model parameters to be tuned and the input variables. A good model should be accurate to capture only variations within the data and avoid noise in the observations as much as possible (overfitting), which requires tuning the model parameters. Simplicity, is another aspect that should be considered, i.e., more input variables in the model could make it more complex without a substantial gain in accuracy. For this reason, two combined techniques: (i) Grid search and (ii) 5 fold cross-validations were performed at different stages of building the model to select the best model. (Discussed following sections).

### 4.1. Selecting the Input Variables

As previously pointed out, the output variables from processing the BSMs include three key variables for each 5-min time step. These are the average travel time ($TT_i$), the average longitudinal acceleration ($Ax_i$) and the average lateral acceleration ($Ay_i$). To check which of these variables is more related to the average travel times to be predicted for the next 5-min step ($TT_{i+1}$), the Pearson's correlation coefficient ($r$) was used. The $r$ was calculated between $TT_{i+1}$ and between each of the following variables: $TT_i$, $Ax$, and $Ay$. The $r$ value pertinent to $TT_i$ was 0.71. The $Ax_i$ and $Ay_i$ showed very low $r$ values of -0.14 and 0.02, respectively. Accordingly, a simple model containing only $TT_i$, excluding $Ax_i$ and $Ay_i$, can still produce accurate results.

### 4.2. Building a Preliminary Model

A simple prediction model was proposed, where only five variables were used to predict the TT one-step ahead as follows:

$$TT_i = f\left(TT_{i-1}, TT_{i-2}, TT_{i-3}, D_{i-1}, H_{i-1,}\right) \tag{5}$$

Where,

  TT $_i$ = Travel time at time step i
  TT $_{i-1}$ = Travel time at time step i-1
  TT $_{i-2}$ = Travel time at time step i-2
  TT $_{i-3}$ = Travel time at time step i-3
  D =Day index (days are indexed 1 to 7 to represent Monday through Sunday)
  H =Hour index (hours are indexed 1 to 15 to represent hours from 5:00 AM to 8:00 PM)

75% of the data were used for tuning the parameters, validating results and training final models. The remaining 25% of the data were used as a final test for evaluating the *XGB* against all proposed tree-based ensemble algorithms.

### 4.3. Performance Measure

The mean absolute percentage error (MAPE) is used for evaluations throughout this paper during both stages: tuning parameters and evaluating the XGB prediction performance, and is given in Equation (6):

$$\text{MAPE} = \frac{1}{N} \sum_{n=1}^{N} \left| \frac{A_i - P_i}{A_i} \right| * 100 \tag{6}$$

Where,

$A_i$ = Ground true value in the i[th] 5 -min interval
$P_i$ = Predicted value in the i[th] 5 -min interval
N = Total number of 5-min intervals in the testing data (= 25%*5400 or 1350 in this study)

### 4.4. Tuning parameters for the Ensemble methods

The ensemble tree-based algorithms proposed in this paper have one or more parameter to tune of the following three main parameters, namely, the number of trees ($t$), learning rate ($L$), and the maximum depth ($d$). The value of ($t$) indicates the number of trees to be averaged, which is a fundamental parameter in all ensemble models. Increasing ($t$) should boost the accuracy and reduce the variance of the prediction model. However, increasing ($t$) to very large values, without considering other parameters $L$ (for Adaboost, GBM, and XGB) and $d$ (for GBM and XGB) might cause overfitting of data, especially for the Adaboost and GBM algorithms. Since the value of ($t$) is proportional to processing speed, it is more efficient, in terms of computation, to pick the smallest value needed to model all variations within the data. The parameter ($L$) is specific only to the Adaboost, GB and XGB algorithms, and shrinks the contribution of each successive tree by the value of ($0 < L <= 1$). As a result, for the Adaboost, GB and XGB there is a tradeoff between the ($t$) and the ($L$). The final accuracy of the model can be boosted by either decreasing ($L$) and increasing ($t$) or vice versa. With such induced shrinkage ($0 < L <= 1$) between successive trees, the model is more regularized and is expected to overcome overfitting and give more accurate results. Finally, the ($d$) parameter is considered a key parameter that affects the performance of the GB and XGB algorithms significantly, and equals to the maximum number of nodes and splits in the tree. The depth of the tree allows for modeling more interactions between the variables input to the model.

The parameters tuning for all models is performed using a five-fold cross-validation technique to search a grid of combinations of all proposed values for the parameters of each algorithm. Grid search is a brute force search through a manually specified subset of the parameters' space of the algorithm. The grid search is guided by the average MAPE of all segments, which is measured by five-fold cross-validation on the training set. Where for each combination of the parameters, 4/5 of the training data is used for training the model and the rest 1/5 of the data is held for calculating the MAPE. The same process is repeated five times using each of the distinctive 1/5 subsamples as the validation data. The five results from the folds are then averaged to produce a single final estimate. Grid search was performed over the following space of the parameters' values:
$t \, \epsilon$ [ 1, 2..., 8000], $d \, \epsilon$ [ 1,2,3,4,5,6,7],
$L \, \epsilon$ [0.0001, 0.0005, 0.001, 0.005, 0.01, 0.05, 0.1, 0.5, 1]

The XGB and GB algorithms were tuned for the optimal combination of ($L$), ($d$) and ($t$), and the Adaboost algorithm was tuned for the optimal combination of ($L$), and ($t$), while the bagging and RF were tuned for the optimal value of ($t$) only. The additional XGB regularization parameters were assigned a constant value of one. Table 1 summarizes the tuned values for the parameters for each of the algorithms. According to the validation results the *XGB* outperformed all of the other models (5.85% MAPE) and at lower computational costs (only 40 trees). The *XGB* low number of trees compared to all other models was expected due to the unique regularization term implemented within the *XGB* algorithm. The GB model also proved its effectiveness in prediction with relatively less accuracy (6.03% MAPE). These results are consistent with the fact that the *XGB* and GB utilize similar base gradient boosting model. As mentioned earlier, the regularization term, memory optimization and cache-line friendly training patterns implemented within the *XGB* are the main reason for the *XGB* outperforming GB in both accuracy and computational speed. For instance, the GB model required 6.1 hours for training the model over the entire grid and 72 seconds for training the best model, unlike the *XGB* model, which needed only 3.2 hours for training the model over the grid and less than 1 second for training the final model. It is worth mentioning that the *XGB* computational speeds can be dramatically increased compared to GB by implementing parallel computing and utilizing multiple cores, other than using one single core as in this study

TABLE 1: SUMMARY OF THE TUNED PARAMETERS FOR THE PROPOSED MODELS

| Model | $t$ | $L$ | $d$ | Average MAPE |
|---|---|---|---|---|
| DT | NA | NA | NA | 7.52% |
| RF | 5000 | NA | NA | 6.61% |
| Bagging | 3000 | NA | NA | 6.60% |
| Adaboost | 3000 | 0.0001 | NA | 6.95% |
| GB | 5000 | 0.001 | 6 | 6.03% |
| *XGB* | 40 | 0.1 | 6 | 5.85% |

### 4.5. Selecting Number of Predictors

This section summarizes the sensitivity analysis performed for selecting the least number of variables input to the model without compromising its performance.

### 1)    Temporal span

The time window (ω), defined as the number of backward time steps included in the prediction model, is an imperative feature. Increasing the value of (ω) improves the model prediction accuracy, at the cost of model complexity and processing time which affects real-time efficiency. To select an optimal value for (ω) while maintaining the model accuracy and simplicity, all the models were re-tuned using 5-fold cross-validation and grid search techniques for six times for different values of ω (ω =1 through 6) and the average MAPE across all the sections was reported for the best model at each value of ω. To eliminate the random variations, the same simulation seed number was maintained in all runs. The results of these runs are depicted in Figure 3. These results show little impact of the time window (ω) for the DT model. For all other models, there was almost no impact beyond time window value of 3. Accordingly, the value of ω = 3 (already implemented) was chosen for further analysis and optimization.

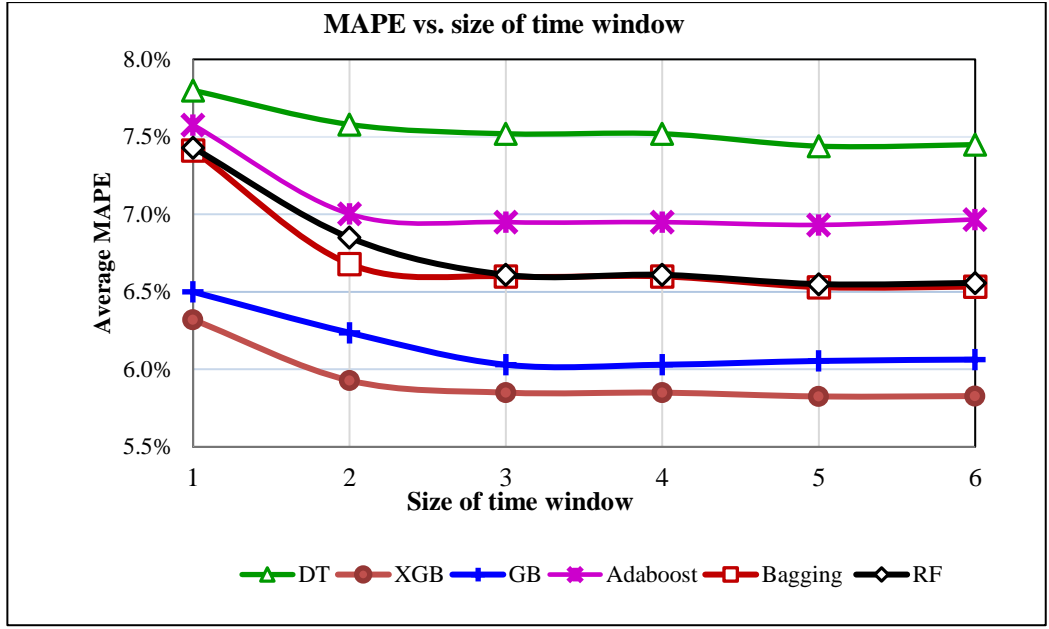

**Figure 3: Sensitivity of Time Window Values (ω)**

*2)    Spatial Range*

In order to capture the impact of spatial patterns accurately during prediction, the travel times of the links upstream and downstream of the link in subject should be considered.   All proposed algorithms were tuned twice using 5-fold cross-validation and grid search techniques, once considering the spatial patterns by including travel times upstream and downstream of the link under study and another time without them. The runs were made for 5 min prediction horizon, and the results are depicted in Figure 4.

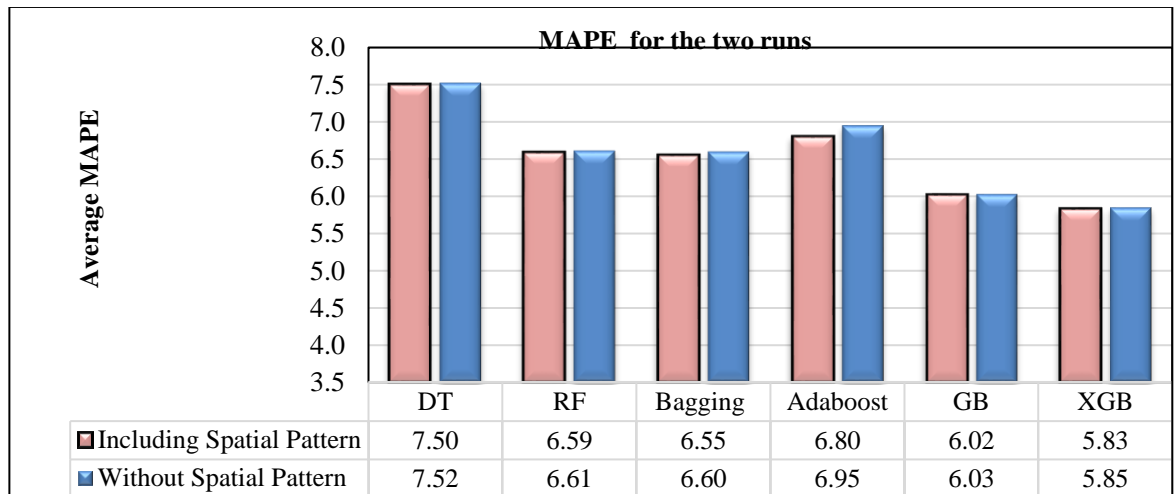

| | DT | RF | Bagging | Adaboost | GB | XGB |
|---|---|---|---|---|---|---|
| Including Spatial Pattern | 7.50 | 6.59 | 6.55 | 6.80 | 6.02 | 5.83 |
| Without Spatial Pattern | 7.52 | 6.61 | 6.60 | 6.95 | 6.03 | 5.85 |

**Figure 4:  Impact of Spatial Pattern On Model**

As shown in the figure, the spatial patterns did not lead to a significant improvement in accuracy. Accordingly, it was decided to choose the simpler model, i.e., the one without spatial patterns. Our models are now ready for predicting travel times only at 5-minute horizon. However, for including higher horizons, the temporal span and spatial range analysis were repeated for each prediction horizon up to 30- minute horizon and the pattern was similar to the 5-minute horizon.

### 4.6. Evaluating the XGB

The ensemble family algorithms were run to predict travel time over the entire stretch for up to 30- minute horizons using the test data. As illustrated in Figure 5, the XGB, GB, Bagging and, RF exhibited robust performance in prediction for long horizons compared to the DT.

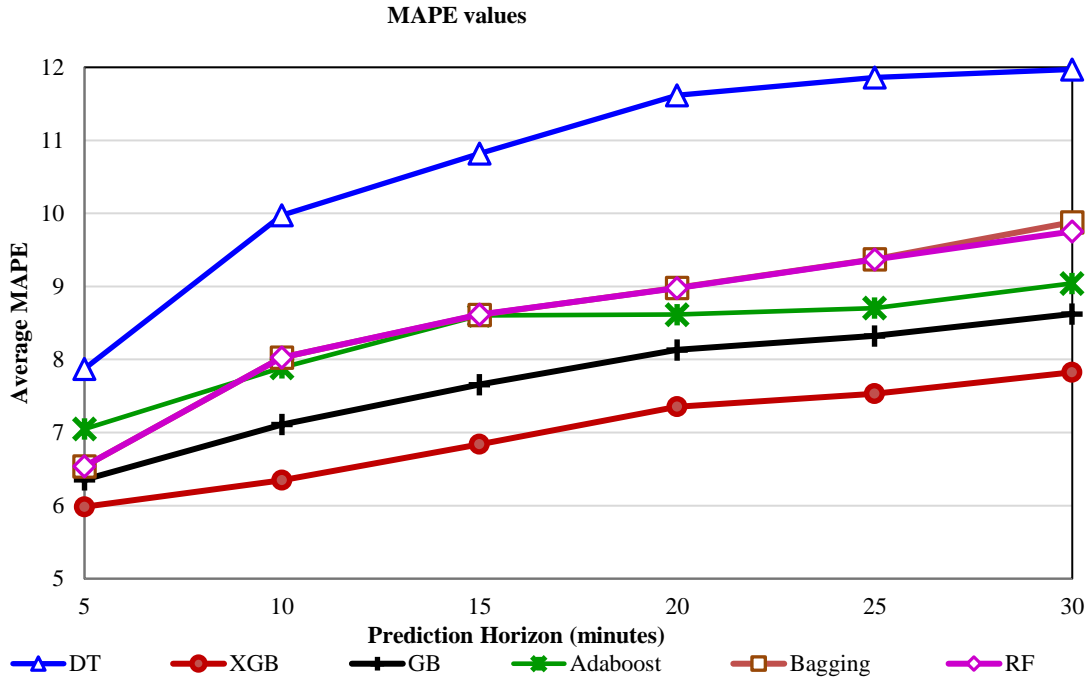

**Figure 5: Comparison of Ensemble Algorithms for Different Prediction Horizons**

According to the figure, the performance of the RF and the Bagging algorithms was almost the same. This is attributed to the low number of variables in the prediction model (only 5 variables). This small number of variables does not allow when using different random subset of features during splitting for each tree for sufficient variation among different trees.

The XGB algorithm outperformed all other algorithms and produced the lowest MAPE at all prediction horizons. The GB also proved its effectiveness in prediction with relatively less accuracy (8.6 % MAPE for 30-min horizon). In addition to being faster than the GB, the XGB algorithm was able to predict travel times at 30-minute horizon with 7.8% MAPE which are very promising results. Moreover, the XGB was used to predict and plot the Heat-Map for each section during short time horizons (5 and 10 minutes) as shown in Figures 6(a) and 6(b) for a sample day. The high prediction accuracy of the XGB is noticeable through the great resemblance between the predicted Heat-Maps and the ground truth Heat-Maps. Furthermore, figures 6(c) and 6(d) illustrate the robustness of the XGB predictions (5 min and 10 min respectively) during the peak periods of the sample day at section 6.

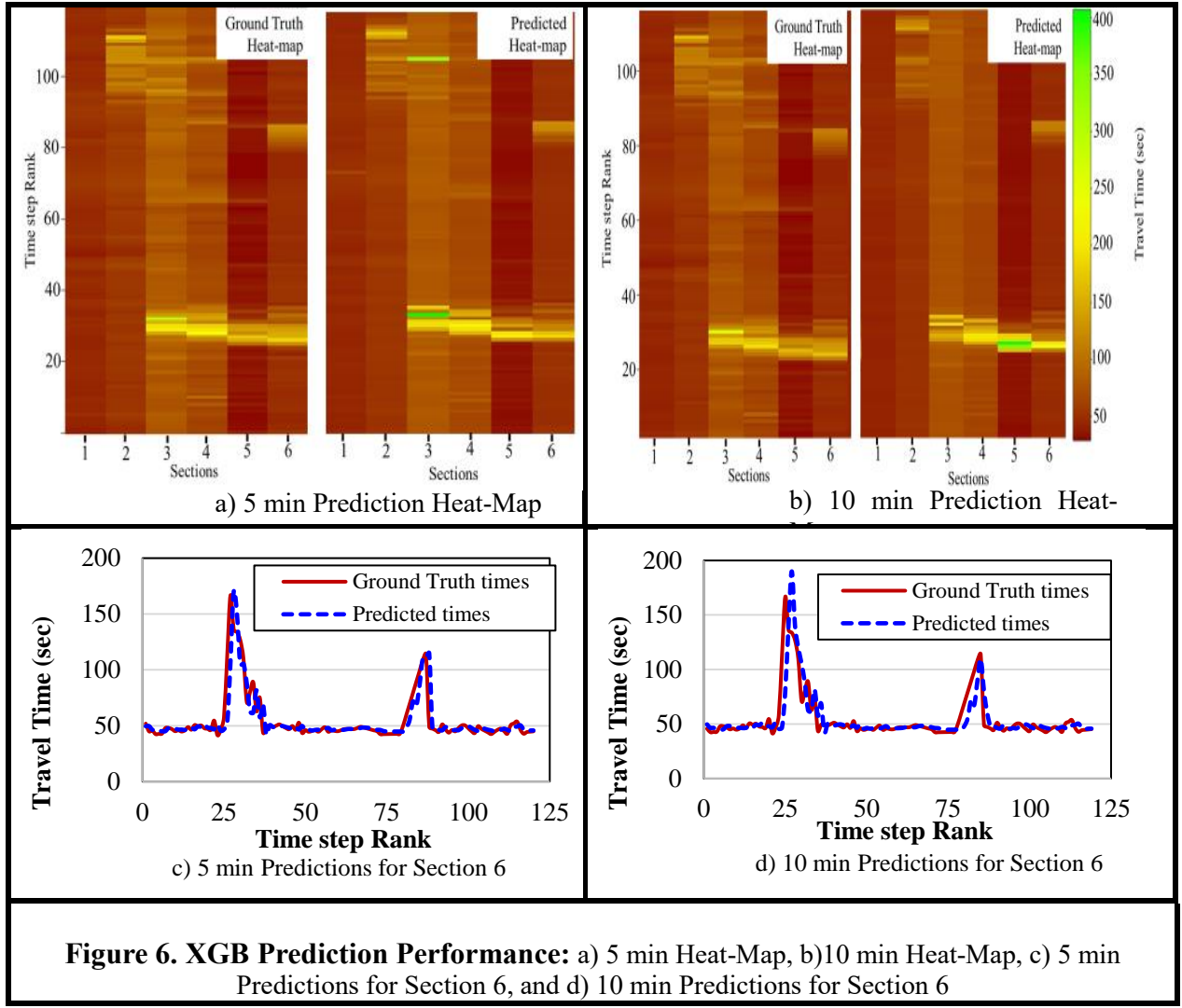

**Figure 6. XGB Prediction Performance:** a) 5 min Heat-Map, b)10 min Heat-Map, c) 5 min Predictions for Section 6, and d) 10 min Predictions for Section 6

## 5. FEATURES IMPORTANCE OF XGB MODEL

A key feature of the XGB algorithm is that it reports the relative importance of, or gain to, the model from each input variable, which is fundamental to the model interpretation, or deciding to exclude unimportant variables. The relative importance (gain) of each variable is the average gain across all trees, and equals the average change in entropy at each tree node the variable was used in splitting. Where the entropy is a metric evaluating the purity at each node (effectiveness of splitting node in improving model classification/regression accuracy). Relative feature importance values were determined for the six sections as depicted in Figure 7.

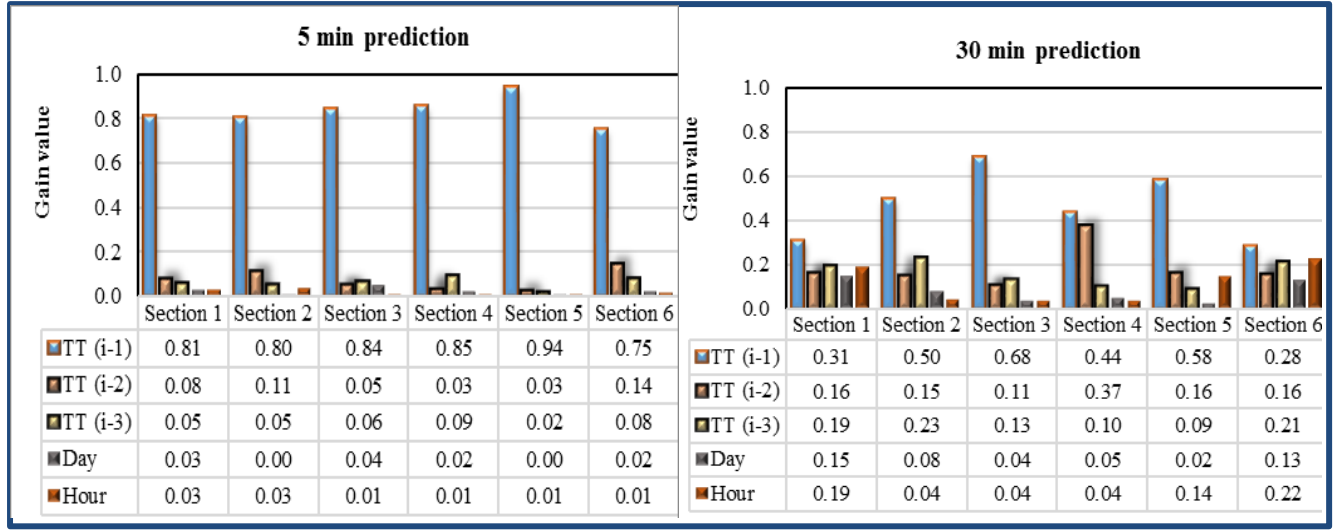

**Figure 7: Relative Importance of Input Variables in XGB Algorithm for Different Prediction Horizons** a) 5 min b) 30 min.

The figure demonstrates the evolution in the behavior of the model as the prediction horizon increases. For 5-min horizon, only travel time of the previous time step, TTi-1, is the most important feature in the model across all sections. This is expected since traffic conditions usually last for more than 5 minutes, and consequently, current travel times are most likely correlated with those of previous 5-min step. When the 30-min horizon is considered, the Day, Hour, TTi-2 TTi-3 parameters, in addition to TTi-1, become more significant in the model. This shows that the model optimizes the required variables for predicting traffic conditions.

## 6. XGB PERFORMANCE UNDER CONGESTED CONDITIONS

To have a better picture of the XGB model's performance under different traffic conditions, a case study was performed to compare the performance of the XGB, GB, Adaboost, Bagging, RF, and DT models in both congested and uncongested conditions, where the congested conditions are the traffic conditions during the peak hours. A 4.7 km long stretch consisting of four links along the eastbound segment of M-14 was used for evaluating the models. The stretch consists of links no. 5, 6, 7, and 9 as shown in Figure 1. These links are referred to as M-14 (A, B, C, and D, respectively). In this case study, the peak periods were 6:00-9:00 AM and 4:00-7:00 PM. Table 2 summarizes the basic statistics for the four segments during both peak and non-peak periods.

TABLE 2: Summary Statistics of the Travel Times during Peak and Non-Peak Periods

| Segment | Length (Km) | Peak | | Non-peak | |
|---------|-------------|------|--------------------|------|--------------------|
| | | Mean | Standard deviation | Mean | Standard deviation |
| M-14A | 1.13 | 69.43 | 51.42 | 41.38 | 21.70 |
| M-14B | 0.98 | 48.99 | 29.24 | 37.39 | 17.79 |
| M-14C | 1.71 | 66.97 | 47.17 | 43.26 | 27.57 |
| M-14D | 0.88 | 28.66 | 3.23 | 27.44 | 1.71 |

Table 3 evaluates the XGB MAPE prediction values against other tree-based models for five-minute intervals over a 30-minute horizon during both peak and non-peak periods. According to Table 3(a), the XGB model outperformed all other models during the peak periods in 23 out of 24 cases. Whereas, in table 3(b), the XGB model outperformed the other models during non-peak periods in 19 out of 24 cases.

In general, the XGB model demonstrated its superior accuracy in predicting travel times during both congested and uncongested conditions. The table indicates that the performance of all models, including the XGB, was almost similar for the links with small variations in travel times (M-14D). For the links with higher variations in travel times (M-14A and M-14C), the higher accuracy of the XGB model compared to the other models, is more apparent, especially during peak periods where the travel time variations are the most.

TABLE 3: EVALUATING THE XGB PERFORMANCE DURING PEAK AND NON-PEAK PERIODS AGAINST OTHER TREE-BASED ALGORITHMS

A) PEAK-PERIODS

| Segment | Model | 5 min | 10 min | 15 min | 20 min | 25 min | 30 min |
|---------|-------|-------|--------|--------|--------|--------|--------|
| **M-14A** | XGB | **8%** | **10%** | **11%** | **11%** | **12%** | **16%** |
| | GB | 9% | 12% | 14% | 16% | 18% | 20% |
| | Adaboost | 14% | 17% | 18% | 18% | 18% | 22% |
| | Bagging | 15% | 19% | 22% | 20% | 22% | 23% |
| | RF | 15% | 18% | 23% | 22% | 23% | 23% |
| | DT | 19% | 24% | 28% | 25% | 35% | 32% |
| **M-14B** | XGB | **6%** | **7%** | **6%** | **7%** | **9%** | **9%** |
| | GB | 8% | 9% | 10% | 10% | 12% | 11% |
| | Adaboost | 8% | 6% | 7% | 10% | 11% | 12% |
| | Bagging | 8% | 8% | 7% | 10% | 11% | 11% |
| | RF | 6% | 9% | 9% | 11% | 12% | 13% |
| | DT | 10% | 14% | 15% | 15% | 16% | 16% |
| **M-14C** | XGB | **10%** | **15%** | **17%** | **18%** | **18%** | **18%** |
| | GB | 12% | 18% | 18% | 18% | 19% | 22% |
| | Adaboost | 15% | 18% | 18% | 20% | 20% | 20% |
| | Bagging | 14% | 18% | 20% | 20% | 19% | 20% |
| | RF | 13% | 18% | 20% | 20% | 21% | 21% |
| | DT | 18% | 22% | 24% | 24% | 25% | 26% |
| **M-14D** | XGB | 2% | **5%** | **5%** | **5%** | **5%** | **6%** |
| | GB | **2%** | 5% | 5% | 5% | 5% | 7% |
| | Adaboost | 2% | 5% | 5% | 5% | 5% | 6% |
| | Bagging | 3% | 5% | 5% | 5% | 5% | 7% |
| | RF | 4% | 5% | 5% | 5% | 5% | 7% |
| | DT | 6% | 7% | 8% | 8% | 8% | 8% |

B) NON-PEAK PERIODS

| Segment | Model | 5 min | 10 min | 15 min | 20 min | 25 min | 30 min |
|---------|-------|-------|--------|--------|--------|--------|--------|
| **M-14A** | XGB | 4% | **5%** | **7%** | **8%** | **7%** | **6%** |
| | GB | **4%** | 6% | 8% | 9% | 8% | 8% |
| | Adaboost | 4% | 6% | 6% | 8% | 7% | 7% |
| | Bagging | 5% | 6% | 8% | 9% | 8% | 8% |
| | RF | 5% | 8% | 8% | 10% | 8% | 8% |
| | DT | 7% | 10% | 11% | 13% | 9% | 12% |
| **M-14B** | XGB | **2%** | **3%** | **4%** | **5%** | **4%** | **4%** |
| | GB | 3% | 3% | 5% | 6% | 5% | 5% |
| | Adaboost | 3% | 3% | 4% | 5% | 4% | 4% |
| | Bagging | 3% | 4% | 5% | 6% | 5% | 5% |
| | RF | 3% | 5% | 5% | 6% | 5% | 5% |
| | DT | 5% | 7% | 8% | 9% | 9% | 9% |
| **M-14C** | XGB | **2%** | **3%** | **3%** | **2%** | **3%** | **3%** |
| | GB | 2% | 2% | 2% | 3% | 3% | 3% |
| | Adaboost | 2% | 3% | 3% | 3% | 3% | 3% |
| | Bagging | 2% | 3% | 3% | 3% | 3% | 3% |
| | RF | 2% | 3% | 3% | 3% | 3% | 3% |
| | DT | 3% | 3% | 3% | 4% | 4% | 4% |
| **M-14D** | XGB | 1% | 1% | **1%** | 2% | 2% | **2%** |
| | GB | **1%** | **1%** | 1% | **1%** | **1%** | 2% |
| | Adaboost | 1% | 1% | 2% | 2% | 2% | 2% |
| | Bagging | 1% | 1% | 1% | 2% | 1% | 2% |
| | RF | 1% | 1% | 1% | 1% | 2% | 2% |
| | DT | 1% | 2% | 2% | 2% | 2% | 2% |

## 7. CONCLUSIONS

Travel time prediction models are a key component of any Traffic Management Center (TMC) aiming for providing travel time predictions. Travel time prediction informs drivers about the anticipated delay along their trip. Such information assists travelers in choosing the least cost route or postponing their departure time to a later time if possible. With the introduction of the Connected Vehicles (CV) technology, more traffic data will be collected and more opportunities will be available for developing more reliable travel time prediction models. Thus, it is imperative to find a prediction model that is able to handle big data and capture complex relations within the data, yet provide insights into the structure of the model. The unique capabilities of the XGB in modeling complex non-linear relationships with high accuracy in addition to providing relative importance of the model input variables make it a promising algorithm for travel time prediction.

This study develops an algorithm for processing the Basic Safety Messages (BSM) across multiple processing nodes and aggregating the results for providing the final spatial-temporal travel time matrix for the entire network at 5 min interval. This travel time matrix was used along with a combination of grid search and five-fold cross-validation techniques for developing the XGB prediction models and tuning its parameters. Unlike the RF and Bagging algorithms that average trees, the XGB grows trees sequentially, which enables modeling complex non-linear relation more accurate. Moreover, unlike the GB and Adaboost models, the XGB algorithm has its unique regularization term that allows the XGB algorithms to model complex non-linear relations without modeling the noise within the data (overfitting).

Moreover, a 9.6 km freeway stretch of US-23 was utilized for evaluating the performance of the eXtreme Gradient Boosting (XGB) algorithm against the most common tree-based ensemble algorithms, and the evaluations are performed at five-minute intervals over a 30-minute horizon. Results show that XGB is superior to all algorithms, followed by the Gradient Boosting (GB). Its travel time predictions from BSM were accurate and consistent with variations during peak periods, with mean absolute percentage error (MAPE) in prediction about 5.9% and 7.8% for 5-minute and 30-minute horizons, respectively. Finally, the XGB algorithm indicated that for the 5-min prediction horizon, only the travel time of the previous time step, $TT_{i-1}$, is the most important feature in the prediction model across all sections. Unlike, the 30-min prediction horizon, where all other variables contribute more in the model, depending on the location of the segment. An additional case study, where the developed models were applied to a 4.7 km stretch along the eastbound of M-14, demonstrated the superior performance of the XGB algorithm to all developed models during congested and uncongested conditions.

When applying the XGB prediction model for a specific time step, all stamps generated during this five-minute duration step will need to be aggregated on a real-time basis. To this end, this study implements a travel time estimation algorithm that is capable of performing the aggregation process simultaneously through parallel processing nodes. This approach simulates the deployment of multiple CV or roadside units in the network for processing data and providing the travel time estimates on a real-time basis. This study presented a first attempt to implement the XGB algorithm in any travel time prediction model. The study also provided a comprehensive evaluation for XGB prediction model against common tree-based ensemble prediction models such as RF, Bagging, Adaboost, and GB.

## REFERENCES

[1] D. Schrank, B. Eisele, and T. Lomax, "2015 URBAN MOBILITY SCORECARD," Texas A&M Transportation Institute and INRIX, Texas A&M Transportation Institute, Aug. 1, 2015.

[2] "Transit Connected Vehicle Research Program." [Online]. Available: http://www.its.dot.gov/factsheets/transit_connectedvehicle.htm. Accessed: Jan. 13, 2017.

[3] N. Lu, N. Cheng, N. Zhang, X. Shen, and J. W. Mark, "Connected vehicles: Solutions and challenges," IEEE Internet of Things Journal, vol. 1, no. 4, pp. 289–299, Aug. 2014.

[4] T. M. Research, "Connected car market - global industry analysis, size, share, growth, trends and forecast 2013 - 2019," 2013. [Online]. Available: http://www.transparencymarketresearch.com/connected-car.html. Accessed: Jan. 13, 2017.

[5] Safety Pilot Model Deployment. Vol. 1542, 2015, pp. 33–36. [Online]. Available: http://www.its-rde.net/data/showds?dataEnvironmentNumber=10018. Accessed: Jan. 13, 2017.

[6] D. Bezzina and J. R. Sayer, "Safety Pilot: Model Deployment Test Conductor Team Report," 2015. [Online]. Available: http://safetypilot.umtri.umich.edu.

[7] M. Vasudevan, D. Negron, M. Feltz, J. Mallette, and K. Wunderlich, "Predicting congestion states from basic safety messages by using big-data graph Analytics," Transportation Research Record: Journal of the Transportation Research Board, vol. 2500, pp. 59–66, Jan. 2015.

[8] E. I. Vlahogianni, M. G. Karlaftis, and J. C. Golias, "Short-term traffic forecasting: Where we are and where we're going," *Transportation Research Part C: Emerging Technologies*, vol. 43, pp. 3–19, Jun. 2014.

[9] E. I. Vlahogianni, J. C. Golias, and M. G. Karlaftis, "Short-term traffic forecasting: Overview of objectives and methods," *Transport Reviews*, vol. 24, no. 5, pp. 533–557, Sep. 2004.

[10] Y. Qi and S. Ishak, "A hidden Markov model for short term prediction of traffic conditions on freeways," *Transportation Research Part C: Emerging Technologies*, vol. 43, pp. 95–111, Jun. 2014.

[11] Elhenawy M. and Rakha H., "Expected Travel Time and Reliability Prediction using Mixture Linear Regression," Accepted for presentation at the *95th Transportation Research Board Annual Meeting*, Washington DC, 2016.

[12] Farokhi, K. S., Masoud, H., & Haghani, A. "Evaluating moving average techniques in short-term travel time prediction using an AVI data set," Presented at *Transportation Research Board Annual Meeting*, Washington, DC 2010.

[13] Y. Qi and S. Ishak, "Stochastic approach for short-term freeway traffic prediction during peak periods," *IEEE Transactions on Intelligent Transportation Systems*, vol. 14, no. 2, pp. 660–672, Jun. 2013.

[14] Ishak, S. and Al-Deek, H., "Performance evaluation of short-term time-series traffic prediction model," *Journal of Transportation Engineering*, vol. 128, pp. 490–498, 2002.

[15] J. Y. Wang, K. I. Wong, and Y. Y. Chen, "Short-term travel time estimation and prediction for long freeway corridor using NN and regression," *2012 15th International IEEE Conference on Intelligent Transportation Systems*, Sep. 2012. Author.

[16] J. W. C. van Lint, S. P. Hoogendoorn, and H. J. van Zuylen, "Accurate freeway travel time prediction with state-space neural networks under missing data," Transportation Research Part C: Emerging Technologies, vol. 13, pp. 347–369, 2005

[17] C. Wu, J. Ho, and D.T. Lee, "Travel-time prediction with support vector regression," *IEEE Transactions on Intelligent Transportation Systems*, vol. 5, pp. 276–281, 2004

[18] L. Vanajakshi and L. R. Rilett, "Support vector machine technique for the short term prediction of travel time," *2007 IEEE Intelligent Vehicles Symposium*, Jun. 2007.

[19] M. Oltean and L. Dioșan, "An autonomous GP-based system for regression and classification problems," Applied Soft Computing, vol. 9, pp. 49–60, 2009.

[20] M. Elhenawy, H. Chen, and H. A. Rakha, "Dynamic travel time prediction using data clustering and genetic programming," *Transportation Research Part C: Emerging Technologies*, vol. 42, pp. 82–98, 2014.

[21] A. Ladino, A. Kibangou, H. Fourati, and C. C. de Wit, "Travel time forecasting from clustered time series via optimal fusion strategy," *2016 European Control Conference (ECC)*, Jun. 2016.

[22] J. Yu, G.-L. Chang, H. W. Ho, and Y. Liu, "Variation based online travel time prediction using clustered neural networks," *2008 11th International IEEE Conference on Intelligent Transportation Systems*, Oct. 2008.

[23] J. Elith, J. R. Leathwick, and T. Hastie, "A working guide to boosted regression trees," *Journal of Animal Ecology*, vol. 77, pp. 802–813, 2008.

[24] Mohammed Elhenawy, Hao Chen and Hesham Rakha (2014), "Random Forest Travel Time Prediction Algorithm using Spatiotemporal Speed Measurements," in 21st World Congress on Intelligent Transportation Systems, Detroit, USA

[25] B. Hamner, "Predicting travel times with context-dependent random forests by modeling local and aggregate traffic flow," *2010 IEEE International Conference on Data Mining Workshops*, Dec. 2010.

[26] Y. Zhang and A. Haghani, "A gradient boosting method to improve travel time prediction," *Transportation Research Part C: Emerging Technologies*, vol. 58, pp. 308–324, Sep. 2015.

[27] "Kaggle Inc," 2017. [Online]. Available: https://www.kaggle.com/. Accessed: Jan. 16, 2017.

[28] "What is the difference between the R gbm (gradient boosting machine) and *XGB*oost (extreme gradient boosting)?," Available: https://www.quora.com/What-is-the-difference-between-the-R-gbm-gradient-boosting-machine-and-*XGB*oost-extreme-gradient-boosting. Accessed: Jul. 16, 2016.

[29] L. Breiman, "Bagging predictors," *Machine Learning*, vol. 24, no. 2, pp. 123–140, Aug. 1996.

[30] A. Mayr, H. Binder, O. Gefeller, and M. Schmid, "The evolution of boosting Algorithms," *Methods of Information in Medicine*, vol. 53, no. 6, pp. 419–427, Aug. 2014.

[31] Tianqi Chen. "Higgs Boson Discovery with Boosted Trees". In: May 2014 (2015), pp. 69–80.

[32] Chen, T. and Guestrin, C. *XGB*oost: A scalable tree boosting system. arXiv preprint arXiv:1603.02754, 2016.

[33] Chen, T. and He, T. *XGB*oost: eXtreme Gradient Boosting. https://github.com/dmlc/*XGB*oost. Accessed: 2015-06-05.